\documentclass[a4paper,10pt]{article}
\usepackage[a4paper]{geometry}
\usepackage[table]{xcolor}

\usepackage[utf8]{inputenc}
\usepackage[T1]{fontenc}
\usepackage[
style=authoryear,
backend=biber,
natbib=true,
maxnames=10,
maxcitenames=2,
uniquelist=minyear,
sorting=nyt
]{biblatex}
\usepackage{subfig}

\usepackage{booktabs}

\usepackage{makecell}
\usepackage[hidelinks]{hyperref}
\usepackage{titlesec}
\usepackage{etoolbox}
\patchcmd{\thebibliography}{\chapter*}{\section*}{}{}
\renewcommand{\cite}{\citet}
\usepackage{orcidlink}
\usepackage{graphicx}
\usepackage{lipsum}
\usepackage{amsbsy,amsmath,amssymb,amsfonts,amssymb}
\usepackage{multirow}

\def\keywords{\vspace{.5em}
{\textbf{Keywords}:\,\relax%
}}

\title{Tracking Semantic Change in Slovene: A Novel Dataset and Optimal Transport-Based Distance}

\author{
Marko Pranjić\orcidlink{0000-0002-8645-9714}$^{\dagger,\ddagger}$
\and
Kaja Dobrovoljc\orcidlink{0000-0002-5909-7965}$^{\dagger,\diamond}$
\and
Senja Pollak\orcidlink{0000-0002-4380-0863}$^\dagger$
\and
Matej Martinc\orcidlink{0000-0002-7384-8112}$^\dagger$
}

\date{%
    ($\dagger$)\;Jožef Stefan Institute, Ljubljana, Slovenia\\%
    ($\ddagger$)\;Jožef Stefan International Postgraduate School, Ljubljana, Slovenia\\
    ($\diamond$)\;University of Ljubljana, Slovenia\\
    [3ex]
}

\begin{document}

\maketitle
\begin{abstract}
\noindent In this paper, we focus on the detection of semantic changes in Slovene, a less resourced Slavic language with two million speakers.
Detecting and tracking semantic changes provides insight into the evolution of language caused by changes in society and culture.
We present the first Slovene dataset for evaluating semantic change detection systems, which contains aggregated semantic change scores for 104 target words obtained from more than 3,000 manually annotated sentence pairs.
We analyze an important class of measures of semantic change metrics based on the Average pairwise distance and identify several limitations. To address these limitations, we propose a novel metric based on regularized optimal transport, which offers a more robust framework for quantifying semantic change. 
We provide a comprehensive evaluation of various existing semantic change detection methods and associated semantic change measures on our dataset. Through empirical testing, we demonstrate that our proposed approach, leveraging regularized optimal transport, achieves either matching or improved performance compared to baseline approaches. 
\end{abstract}


\section{Introduction}

Language is a dynamic system that reflects the cultural or technological development of society~\citep{aitchison2001language}. This means that the meaning of a word is constantly changing and evolving through use in social interactions and changes in cultural practices. The change in meaning of a word usually goes through several polysemous stages \citep{hopper1991some}, which makes the identification and understanding of these changes notoriously difficult. Nonetheless, work on this topic is important for linguistic research and social analysis, as changes in language reflect changes in society and can be used as a proxy for the detection of cultural and linguistic trends \citep{GillaniDynamicPerception19}. The detection of these changes can also be used to improve many natural language processing (NLP) tasks. For example, it could improve the temporal sensitivity of current contextual language models, which are currently static and cannot adapt to changes over time \citep{rosin2022time}.

The field of lexical semantic change detection is a very active area of research. While the first systems for automatic change detection were developed more than a decade ago, research on this topic gained momentum with the idea of using embeddings for the construction of temporal representations \citep{Gulordava,hamilton2016cultural}. This led to the development of several state-of-the-art semantic change detection systems and the need for manually labeled gold standard datasets of high quality for training and evaluating these systems. To address the problem of the lack of gold standard and the lack of standardization in terms of evaluation tasks and datasets, several shared tasks on this topic have recently been organized \citep{schlechtweg2020semeval,Basile2020DIACRItaE,kutuzov2021rushifteval,d-zamora-reina-etal-2022-black,fedorova-etal-2024-axolotl24}.

These shared tasks mostly covered lexical semantic change detection in high-resourced languages with many speakers, such as English, Russian, German, Italian, and Spanish. Although some less-resourced languages, such as Latin \citep{schlechtweg2020semeval}, were also considered, most less-resourced languages still lack evaluation datasets that could be used for the evaluation of lexical semantic change detection systems. In the absence of gold standards available to the community, this hinders progress and leads to a widening of the already large gap in NLP research between low-resourced and high-resourced languages.

To address this problem, in this paper we present a gold standard semantic change detection dataset for Slovene, a language with about two million speakers and with significantly fewer resources than high-resourced languages, such as English \citep{elrcwhitepaper}. The dataset covers two time periods with a 20-year gap in between, extracted from the Slovene reference corpus Gigafida 2.0 \citep{krek2020gigafida}.
The first period includes fiction, textbooks, and news texts from 1990 to 1997, while the second contains similar genres from 2018. It contains manually annotated semantic relatedness scores for 3,150 sentence pairs and aggregated semantic change scores for 104\footnote{Annotation was carried out on 105 words, but one word was later excluded due to identified issues in the preprocessing.} words.
Both the annotated semantic relatedness scores for sentence pairs as well as aggregated semantic change scores are publicly available.\footnote{The dataset is available at: \url{http://hdl.handle.net/11356/1651}}

We conducted an extensive evaluation on our dataset using various systems and metrics\footnote{Throughout
this study, we employ the terms distance and metric broadly to describe measures of semantic change. While these formulations do not necessarily satisfy the mathematical properties of a true metric, they
remain useful tools for quantifying linguistic shifts across time periods.} for quantifying semantic shift between two time periods. Among these metrics, we identified a limitation in one of the top-performing metrics and proposed an enhanced approach using regularized optimal transport to address its shortcomings. 
More precisely, we measure the distance between collections of semantic vectors through the concept of transport cost.
Our experiments demonstrate promising performance on a novel dataset.\\

\noindent The main contributions of this work are as follows:
\begin{itemize}
    \item A construction of the first manually labeled gold standard semantic change detection dataset for Slovene, a low-resourced language with about two million speakers.
    \item A thorough analysis of the constructed dataset, the annotation effort and the resulting annotations.
    \item We show that a number of functions used to detect semantic change have a simpler form and identify some of their limitations.
    \item A novel measure of semantic change based on regularized optimal transport that improves evaluation results compared to several embedding-based semantic change detection systems on the new dataset. 
    \item The first comparative evaluation of multiple semantic change detection systems in Slovene.

\end{itemize}

The paper is structured as follows -- in Section \ref{sec:related_work} we discuss related work on semantic change detection including approaches and the metrics to quantify the semantic change. In Section \ref{sec:dataset} we describe the construction of the dataset, discuss the annotations procedure, and analyze agreement between annotators. In Section \ref{sec:methodology} we analyze Average pairwise distance metric and describe our newly proposed approach for semantic change detection using optimal transport. The experimental setup is discussed in Section \ref{sec:experimental_setup} and the results of the evaluation experiments are presented in Section \ref{sec:evaluation}. Finally, in Section \ref{sec:conclusion} we draw conclusions  and provide suggestions for further work.

\section{Related Work}
\label{sec:related_work}

The study of semantic change has occupied scholars long before modern linguistics emerged in the late 19th and early 20th centuries and introduced new methodologies to the study of language change, with earliest references assigned to \cite{008903202} and even Aristotle \citep{tahmasebi2021computational}.
Automatic detection of temporal semantic change has recently become increasingly important, as it is useful not only in linguistics and lexicography, but also in various text stream monitoring tasks, such as event detection \citep{kutuzov2017tracing}, viewpoint analysis~\citep{azarbonyad2017words,martinc2021embeddia}, news stream analysis \citep{montariol2021scalable}, or the monitoring of discourse changes during crisis events~\citep{stewart2017measuring}. Several studies and shared tasks have recently been published on this topic \citep{schlechtweg2020semeval,Basile2020DIACRItaE,kutuzov2021rushifteval,d-zamora-reina-etal-2022-black,fedorova-etal-2024-axolotl24} to promote the development of new systems and compare different approaches.

\subsection{Evaluation of Semantic Change}
The evaluation of this task remains a challenge as it relies on manually annotated gold standard corpora covering multiple word usages, which are still scarce. The first gold standard dataset for the evaluation of semantic change systems that we are aware of, is the dataset created by \cite{Gulordava}. It contains English words from two time periods, the 1960s and 1990s. The annotators determined the degree of semantic change score for each word based on their intuition, i.e. without looking at the context. This procedure was later considered problematic, as an annotator might forget or not know a particular meaning of the word \citep{montariol2021scalable}.

For this reason, the procedure most commonly used in the creation of recent corpora for assessing semantic change is to (1) annotate pairs of sentences in which the target words have either the same or a different usage (i.e. the procedure first used in the creation of word sense disambiguation datasets \citep{erk2013measuring}) and (2) aggregate these pairwise annotations into semantic shift scores. The most common method for this is the COMPARE metric \citep{schlechtweg2018diachronic}, which is used, for example, in \cite{kutuzov-pivovarova-2021-three,rodina-kutuzov-2020-rusemshift} and also in this research. While the COMPARE metric represents a simple average of pairwise relatedness scores, \cite{schlechtweg2020semeval} employs a more sophisticated approach by clustering a diachronic word usage graph to derive word sense distributions across different time periods. Although this method is more complex, it has been shown empirically to converge with the COMPARE metric, as demonstrated in \cite[p. 64]{Schlechtweg2023measurement}.

When it comes to number of time periods and the number of words in the manually annotated semantic shift datasets, most of them contain only two time periods \citep{Gulordava,schlechtweg2020semeval,schlechtweg2018diachronic,Basile2020DIACRItaE,d-zamora-reina-etal-2022-black} and about 100 target words or less, which is due to the time-consuming manual annotations. To our knowledge, the Russian RuShiftEval \citep{kutuzov2021rushifteval} and RuSemShift \citep{rodina-kutuzov-2020-rusemshift} are the only datasets that cover three distinct time periods: pre-Soviet, Soviet, and post-Soviet. While RuSemShift focuses on two specific word sets -- one tracking changes between the pre-Soviet and Soviet periods, and another between the Soviet and post-Soviet periods -- RuShiftEval uses a unified word set to systematically track words across all three time periods. In terms of the number of words, RuShiftEval is also the largest with 111 target words.
It is closely followed by the English dataset proposed by~\cite{Gulordava}, the Russian dataset created by \cite{rodina-kutuzov-2020-rusemshift}, and the Spanish dataset from \cite{d-zamora-reina-etal-2022-black}, all consisting of about 100 words. In contrast, the datasets created for four languages (English, German, Latin, Swedish) as part of SemEval-2020 Task 1: Unsupervised Lexical Semantic Change Detection \citep{schlechtweg2020semeval} are smaller and contain between 30 and 50 words each. The smallest are an Italian dataset from EVALITA 2020 task: Diachronic lexical semantics in Italian (DIACR-Ita) by \cite{Basile2020DIACRItaE}, and German DURel dataset \citep{schlechtweg2018diachronic}, which contain only 23 and 22 words, respectively. An overview of several comparable datasets with available languages and number of annotated words is provided in Table \ref{tbl:dataset_comparison}.
\begin{table}[ht]
\begin{tabular}{llc}

\textbf{Dataset} & \textbf{Language} & \textbf{Number of words} \\
\hline
\multirow{4}{21em}{SemEval2020: Task1 \citep{schlechtweg2020semeval}}
& ENG & 37 \\
& GER & 48 \\
& LAT & 40 \\
& SWE & 31 \\
DURel \citep{schlechtweg2018diachronic} & GER & 22 \\
DIACR-Ita \citep{Basile2020DIACRItaE} & ITA & 23 \\
Google Books Ngram \citep{Gulordava} & ENG & 100 \\
RuSemShift \citep{rodina-kutuzov-2020-rusemshift} & RUS & 100 \\
LSCDiscovery \citep{d-zamora-reina-etal-2022-black} & SPA & 100 \\
RuShiftEval \citep{kutuzov2021rushifteval} & RUS & 111$^a$ \\
Slovene Sem. Change Dataset \citep{11356/1651} (ours) & SLO & 104 \\
\hline

\end{tabular}
$^a$~\footnotesize{ {RuShiftEval dataset contains annotations from three time periods, in effect annotating three times more word comparisons.}}
\caption{Size comparison with several comparable semantic change detection datasets with available languages and number of words in the dataset.
}
\label{tbl:dataset_comparison}
\end{table}

In order to aid with the annotation efforts, the DURel Annotation Tool\footnote{Available at: \url{https://durel.ims.uni-stuttgart.de/}}  has been developed and made freely available. This tool generates a Word Usage Graph (WUG), where nodes represent individual word usages and edges between them denote the semantic relatedness of these usages. The repository of WUGs \citep{wugs-online} includes resources from nine languages, with some of the graphs further enriched with sense definitions through the work of \cite{kutuzov-etal-2024-enriching}.

\subsection{Systems for Automatic Change Detection}
The first systems for automatically detecting semantic change were developed more than a decade ago. They relied on frequency-based methods~\citep{juola2003time,hilpert2008assessing}, which are rarely used today, as the invention of word embedding representations \citep{mikolov2013distributed} led to the development of more effective methods for this task. A detailed overview of these slightly older frequency and embedding-based methods can be found in \cite{tahmasebi2018survey,SOTAKutuzov,tang_2018}.

Current work on semantic change detection employs two distinct unsupervised methodologies. They are based on the construction of temporal representations using either static or contextual embeddings. The methodology employing static embeddings is based on training a static embedding model for each temporal segment of the corpus and then aligning these models to make them comparable. This can be achieved by using \textit{incremental updating}~\citep{kim2014}, where an embedding model is trained from scratch on the first time slice of the corpus and then updated at each subsequent time slice. Another option is to use \textit{vector space alignment}~\citep{Hamilton2016}. Here, the embedding models are trained independently for each time slice and at the end an alignment is performed by optimizing the geometric transformation. Another approach, that relies on static embedding models, is based on the comparison of target word neighbors (i.e. words with embedding representations that are very similar to the target word representations) in different time slices~\citep{hamilton2016cultural,yin2018global,gonen-etal-2020-simple}. In \cite{gonen-etal-2020-simple}, for example, they use static embeddings to obtain period-specific representations. In each time period, a word is represented by its top nearest neighbors according to the cosine distance and the semantic change is measured as the size of the intersection between the lists of nearest neighbors of two periods.

All methods using static embedding models suffer from limitations in sensitivity and interpretability due to the fact that each word has only one representation within a time slice.\footnote{One exception is the system from~\cite{frermann-lapata-2016-bayesian}, which employs a Bayesian model to track the smooth and gradual evolution of a set of senses for each target word.}
These limitations can be mitigated by using contextual embeddings such as BERT~\citep{Devlin2019BERTPO}, where a different embedding representation is generated for each context in which the word occurs, and enables modeling of the word polysemy.

Another interesting approach, also using contextual embeddings, was proposed by \cite{rosin2022time}. To make a BERT-like model sensitive to time-specific word usage, they propose to concatenate a special time token to each text sequence during the finetuning of the masked language model, thus directly incorporating temporal information into the training process. The language model finetuned in this way is able to predict the time period of each sentence and detect semantic change by looking at the distribution of the predicted periods for each temporal segment (e.g. a uniform distribution indicates no semantic change, while a non-uniform distribution implies a change).
Beside finetuning on the temporal token classification, \cite{zhou-etal-2023-finer} evaluate the performance of the semantic change detection with regards to finetuning on other NLP tasks. The results show that finetuning on any single NLP task like Grammatical Error Correction, Part-of-Speech tagging or Natural Language Inference worsens the performance of the model on the semantic change detection task. But this trend reverses if the word representation is taken to be an average of contextual representations independently finetuned on several tasks. Even though the finetuning step itself can be replaced with Transformer Adapters \citep{pfeiffer2020AdapterHub}, the downside of combining different combinations is that it is unclear if a particular finetuning task will lead to better performance in combination with other tasks or will, in fact, decrease the overall performance \citep[see][]{zhou-etal-2023-finer}.

Further research of different word representations led to the recent result from \cite{card-2023-substitution}, where an improvement in semantic change performance was achieved by representing a word with a set of $k$ most likely replacements from a masked language model. Such replacements are aggregated for each target word across the whole time slice to a single distribution of top-$k$ replacements and JSD is used to measure the difference between such time-slice replacement distributions. Advantage of this approach is that dominant semantic meaning for the time slice naturally emerges as the most likely replacement, thus providing a level of explainability. A downside of the approach is that the performance improvement is not consistent across different datasets. Although there is no explicit clustering step in this approach, the resulting list of replacements does represent a cluster of most likely words.

A related task in semantic disambiguation is the Word-in-Context (WiC) task. It is formulated as a binary classification problem, where the goal is to determine whether two instances of the same word appearing in different contexts share the same semantic meaning. The data for this task often contains separate train and test splits, like in \cite{pilehvar-camacho-collados-2019-wic}, and shared task competitions \citep{armendariz-etal-2020-semeval,martelli-etal-2021-mclwic}. It can be framed as both a supervised and an unsupervised problem and often contains a part of the dataset for training the prediction model.
A specific feature of systems for WiC task is a binary classification layer that predicts whether semantic usage of a word is related. In \cite{homskiy-arefyev-2022-black}, two sentences, representing two usages of the target word, are presented as input to the XLM-R \citep{conneau-etal-2020-unsupervised} model. Target word representations are derived from the final model layer for both of those sentences and concatenated before a classification layer.
Somewhat more involved approach is described in \cite{cassotti-etal-2023-xl}, where the target word in a sentence is surrounded with a pair of tokens (\texttt{<t>}, and \texttt{<{\textbackslash}t>}). In this case, the model receives a single sentence as an input and the model is trained as a Siamese neural network, similar to the approach taken by the SentenceBERT \citep{Reimers2019SentenceBERTSE} model.
In contrast to the WiC task, this work is more focused on a graded change detection where a list of words is ordered based on the measure of semantic change. While the dataset introduced in this work can be adapted to a WiC evaluation resource by selecting an appropriate threshold for the annotated differences, its limited number of target words (104) and the inclusion of detailed information about the scale of semantic differences make it more suitable for evaluating systems for graded semantic change.

\subsection{Measuring Semantic Change}

The related work on graded semantic change discussed above primarily focuses on word representations, exploring various research directions in developing those semantic representations.
Functions used to measure semantic change between such representations are typically adapted from prior studies.
When multiple distance measures are used in evaluation, like in the studies by \cite{kutuzov2020uio,montariol2021scalable,zhou-etal-2023-finer}, the results frequently exhibit inconsistencies as superior performance achieved using one measure of semantic change does not guarantee comparable improvement measured with another one.

We can make a distinction between measures of semantic change and divide them to distribution-based and prototype-based approaches \citep[cf.][]{periti-24-survey}.

\textbf{Distribution-based} approaches first estimate a probability distribution of word senses or word replacements separately for periods $T_1$ and $T_2$, and measure differences between those distributions.
The most common approach to estimate the distribution uses clustering, like performed in \cite{addMoreClusters2020}.
Given the vector representations from two time periods ($T_1$ and $T_2$), clustering-based approaches rely on unsupervised clustering of all available representations ($T_1 \cup T_2$) of a target word to a predefined number of clusters. The idea is that clustering will capture different senses of the target word and proportion of items assigned to each cluster can be used as a distribution of word senses in a given period. Measure of semantic shift for a target word is reflected through the measure of difference between word sense distributions within $T_1$ and $T_2$.
Some alternatives to clustering are estimating distribution of most likely replacements for a target word using masked language model as done in \cite{card-2023-substitution}, or estimating a distribution of predicted time-period of a sentence, like in \cite{rosin2022time}.

The most popular method for comparing the derived distributions from different time periods is the Jensen-Shannon divergence (JSD) \citep{lin1991divergence}, which is used in several studies, like \cite{giulianelli-etal-2020-analysing,addMoreClusters2020,giulianelli-etal-2022-fire,laicher-etal-2021-explaining}. In \cite{montariol2021scalable}, they propose to replace JSD with Wasserstein distance (WD) \citep{Kantorovich-ot}, which was motivated by the observation that WD, in addition to comparing cluster assignment distributions, also takes into account distances between clusters in semantic space, leading to better performance.

Alternatively, in the \textbf{prototype-based} \citep{rodina-kutuzov-2020-rusemshift,10.1007/978-3-030-72610-2_13,periti-etal-2022-done,} approaches, all vectors from a single time period are averaged to derive a period-specific word representation. Measure of semantic shift in this case simplifies to comparing average word representations across periods. 
Simplest comparison of such averaged representations is performed with a cosine distance (COS), used, for example, in \cite{beck-2020-diasense,addMoreClusters2020,rosin2022time}.
Some other measures using word prototypes are Inverted cosine similarity over word prototypes (ICOS-PRT) from \cite{kutuzov2020uio}, and the Distance between prototype embeddings (PDIS) introduced in \cite{periti-etal-2022-done}. Instead of averaging all period-specific vectors to a single word prototype, PDIS uses clustering in order to derive word senses and derives a measure of semantic change using several sense-specific prototypes for every target word.

In \cite{giulianelli-etal-2020-analysing}, it was argued against aggregation-based approaches that average all word senses in a single representation and they proposed using Average pairwise distance (APD).
They note that this metric, based on individual distances between all pairs across time periods, also avoids errors stemming from clustering process. \cite{periti-24-survey} describe APD as a metric that does not require clustering nor averaging. As a function of semantic change that removes important limitations, it was used to derive several other measures, like a Difference between token embedding diversities (DIV) from \cite{kutuzov2020distributional}
and a word \textit{variation coefficient} used in \cite{addMoreClusters2020}. A variant of APD was used in \cite{laicher-etal-2021-explaining} to measure the average period-specific APD score (APD-OLD/NEW). In \cite{giulianelli-etal-2022-fire}, they use an average of COS and APD (APD-PRT) to improve results on several languages. Similar to the usage of sense-specific prototypes in PDIS, there is also a comparable variant of APD introduced in \cite{kashleva-etal-2022-black} that uses distances between pairs of senses (cluster centroids) instead of all word representations.

In related work, theoretical analysis of used semantic change measures is lacking and they are compared only empirically. Among them, APD is considered a strong candidate with good results and a number of variants that further improve the results on specific data \citep[cf.][]{periti-24-survey}.
In this paper, we show that the above description of APD as a measure that avoids aggregation is incorrect and that it is a prototype-based metric. We show a much simpler expression to calculate APD that makes the averaging of period specific vectors inside the APD calculation obvious and simplifies APD to a dot product of those average vectors. This simplification helps in understanding of APD and its variants, highlights some limitations, and enables much faster implementation.

In order to address identified limitations, we propose using a distance based on Optimal transport (OT). As OT approach was previously also used for WD metric, we consider our approach as an unification of WD and APD metrics.  We show, how the regularization of the optimal transport plan based on \cite{NEURIPS2019_159c1ffe} produces results comparable to the APD and show that APD is a special case of regularized OT, where the regularization term completely overrides the contribution of the optimal transport plan. In this light, regularized OT can be regarded as a weighted-APD metric with weights placed on pairwise distances in a way that pairs on the optimal transport path contribute more to the total distance.  

The optimal transport problem \citep{monge1781memoire} has proven to be a valuable tool in many natural language processing applications, such as training of text generation models \citep{li-etal-2020-improving-text}, the matching of interpretable text features \citep{swanson-etal-2020-rationalizing} or the optimal choice of model vocabulary \citep{xu-etal-2021-vocabulary}. It is often used to derive some kind of distance on the data. In the Word movers distance \citep{pmlr-v37-kusnerb15}, optimal transport is used to model the distance between text documents represented as a set of trained word vector representations. The MoverScore \citep{zhao-etal-2019-moverscore}, a scoring metric for text generation with contextualized embeddings, uses optimal transport to find an optimal global alignment of tokens. \cite{lee-etal-2022-toward} uses optimal transport to provide interpretable semantic text similarity. In the context of semantic change detection, \cite{montariol2021scalable} applies optimal transport to detect the semantic change, but unlike our work, their approach focuses on measuring the distances between word cluster centroids, while we completely avoid the clustering step, give equal weight to each usage example, and allow for finer-grained differences.

\subsection{Semantic Change Detection in Slovene}
For Slovene, the automatic semantic change detection methods have not yet been systematically evaluated, as no evaluation data was available. However, there has been some related work dealing with computer-aided analysis of language change, mainly in the field of computer-mediated communication. In \cite{pollak-et-al-2019}, a comparative collocation extraction based on statistical measures was used to identify collocates typical of Slovene computer-mediated communication, while in \cite{fiser2019distributional} embeddings to extract semantic change candidates from Slovene Twitter data was used. Similarly, in \cite{gantar2018leksikalne}, the Slovene reference corpus of computer-mediated communication was leveraged to identify changes in vocabulary and meaning through collocation analysis.
In \cite{martinc2021embeddia}, a semantic analysis of several English-Slovene word pairs related to immigration was performed and found both shared and distinct patterns in semantic correlations.

\section{Dataset Construction}
\label{sec:dataset}

\subsection{Corpus Selection}
\label{sec:corpus-selection}
To construct a temporal corpus for evaluating semantic change detection models, we extracted texts from Gigafida 2.0 \citep{krek2020gigafida}, the reference corpus for written standard Slovene. This corpus was selected due to its extensive size (over 1 billion words) and its coverage of texts spanning from 1990 to 2018. The corpus includes a diverse range of text types, such as fiction, textbooks, and news. Notably, newspapers account for nearly half of the corpus, online content for approximately a quarter, and magazines contribute about one-sixth of the total word count. 

From this corpus, we extracted two different temporal sections with the largest possible temporal gap between them, which ensures an identifiable and significant semantic change between several words. In order to obtain temporal chunks of sufficient size, we decided that each chunk should cover at least one year. The first chunk consists of texts from 2018, the final year covered by Gigafida 2.0, and contains about 80 million words. Initially, we considered using texts from 1990 for the second chunk to maximize temporal distance. However, due to insufficient text data from Gigafida 2.0 in 1990 alone, we extended the time period to include texts from 1990 to 1997, resulting in a total of approximately 70 million words. 
Detailed statistics about the sources for the data in each time period of the dataset are provided in Table \ref{tbl:datasets}.

\begin{table}[!ht]
\centering
\caption{Dataset statistics at the level of different time periods (1990--1997 and 2018) and sources. Note that only sources with five or more documents have been included in the table. Therefore, the number of documents and the number of words do not match with the numbers given in the rows ``All''.}
\label{tbl:datasets}
\begin{tabular}{lrr}
\hline
\textbf{Sources from 1990 to 1997}  & \textbf{Num. docs} & \textbf{Num. words} \\
\hline
Dolenjski list & 1776 & 9,540,845\\
Novi tednik & 1599 & 1,048,747\\
Tehni\v ska zalo\v zba Slovenije & 1159 & 2,256,914\\
Dr\v zavni zbor Republike Slovenije & 630 & 1,387,402\\
Urbar & 392 & 336,617\\
Dnevnik & 374 & 17,449,311\\
Mladina & 339 & 13,263,063\\
neznani zalo\v znik & 155 & 1,117,711\\
DZS & 112 & 5,313,083\\
Gorenjski glas & 84 & 4,093,257\\
Zgodovinsko dru\v stvo za Ju\v zno Primorsko & 63 & 637,040\\
Dru\v stvo izdajateljev \v casnika 2000 & 40 & 757,399\\
Krka zdravili\v s\v ca & 27 & 180,655\\
Radio-Tednik & 23 & 4,769,337\\
Infomediji & 21 & 2,004,767\\
\v Studentska organizacija Univerze, \v Studentska zalo\v zba & 14 & 590,129\\
Cistercijanska opatija Sti\v cna & 13 & 1,081,784\\
Kmetijska zalo\v zba & 12 & 280,823\\
Zavod RS za \v solstvo & 8 & 359,655\\
Zveza geografskih dru\v stev Slovenije & 8 & 107,697\\
Sidarta & 7 & 317,158\\
Delo & 6 & 70,966\\
Klub \v studentov MF & 6 & 3,717\\
Desk & 6 & 840,479\\
Karantanija & 6 & 244,399\\
Dru\v stvo 2000 & 5 & 107,584\\
\textbf{All} & 6939 & 69,794,466\\
\hline
\textbf{Sources from 2018}  & \textbf{Num. docs} & \textbf{Num. words} \\
\hline
sta.si & 260 & 24,263,826\\
rtvslo.si & 231 & 24,289,885\\
siol.net & 102 & 10,561,982\\
delo.si & 87 & 7,228,310\\
svet24.si & 86 & 8,185,470\\
dnevnik.si & 53 & 5,147,473\\
24ur.com & 21 & 1,631,404\\
slovenskenovice.si & 19 & 1,162,930\\
Litera & 5 & 273,111\\
\textbf{All} & 870 & 83,111,440\\
\end{tabular}
\end{table}

\subsection{Word List Creation}

When creating the target word dataset, we had to ensure that the following conditions were met:

\begin{itemize}
 \item The dataset contains words whose usage has changed between two time periods.
 \item The dataset contains words whose usage remains constant in both time periods.
 \item Since the usage of words is usually gradual, i.e. it hardly ever changes directly from one usage to another, but typically goes through several polysemous phases \citep{hopper1991some}, the dataset should contain several polysemous words that are in different phases of usage change. This would allow us to measure the degree of change between two different time periods, rather than just making a binary decision about whether the word has changed or not.
 \item The change/consistency in the usage of a particular selected target word should be reflected in the temporal dataset we create.
\end{itemize}

To fulfill the above conditions, we proceeded in much the same way as in \cite{rodina-kutuzov-2020-rusemshift,schlechtweg2020semeval,kutuzov2021rushifteval} by first selecting changed words and then supplementing the list of target words with filler words, i.e. random words that have a similar frequency distribution in both time periods of the dataset. To find changed words, we first searched for related work on this topic and were able to find a list of words with identified and labeled semantic changes created as part of a study by \cite{gantar2018leksikalne}. Unfortunately, we found that this list mainly contains neologisms, e.g. ``\textit{mi\v ska}'' (eng. ``mouse''), which is on the list due to the invention of the computer device), and informal slang words, e.g., ``\textit{nor}'' (eng. ``crazy''), which is on the list due to its informal use as a synonym for ``very good'', that were not used in the constructed dataset in a meaning specified by the list or were very rare. For this reason and because a completely manual search of our large temporal dataset for words experiencing temporal usage change was infeasible due to time and cost constraints, we decided to find a manageable number of candidate words for manual inspection with the help of existing semantic change detection systems.

We opted to use three automatic change detection systems to obtain a large set of candidate words that were afterwards manually inspected. By using three systems instead of just one, we tried to minimize the influence of inductive bias of each specific system on the candidate selection process and maximize the diversity of the candidate word set. Each system retrieved 1,000 candidate words which, according to the system, changed its usage the most between the two time periods in the corpus. An additional constraint was that each retrieved word should occur at least 30 times in each temporal segment, which meant that the entire vocabulary we considered shrank to roughly 50,000 words. The candidate words were ranked according to the perceived semantic change by each system, with rank 1 being assigned to the most changed word. The threshold of 1,000 was selected as an optimal trade-off between the diversity of obtained candidates and the practicality of manual inspection. It also represents the upper limit of the number of candidates that can be feasibly inspected within the constraints of our budget and time. 
 
The systems used were the clustering method proposed by \cite{montariol2021scalable}, the {\em Nearest Neighbors} method proposed by \cite{gonen-etal-2020-simple}, and the \textit{SGNS+OP+CD} method proposed by \cite{hamilton2016cultural}, which refers to a semantic change detection method using static word embeddings --- we apply the Skip-Gram with Negative Sampling (SGNS) model independently to two periods, align the embeddings using Orthogonal Procrustes (OP) and use Cosine Distance (CD) to compute the semantic change. This model has shown competitive performance in the SemEval-2020 Task 1 \citep{schlechtweg2020semeval}, outperforming systems based on contextual embeddings on several corpora. Since each of these models provided a list of words ordered by usage change, this allowed us to compute an average rank across the three systems for each word in the vocabulary. After that, two native speakers manually checked the final list of 1,000 candidate words ordered by the average rank. After deduplication and filtering (e.g., we manually removed several corpus artefacts), they manually selected 49 words. 

The manual selection criteria followed two main principles, i.e., semantic and evolutionary diversity. To appease semantic diversity, the manual selection tried to maximize the number of topics covered by the main senses of the selected words. For this reason, several highly ranked words belonging to the over-represented topics were not chosen. An example of this would be the word ``diagonalen (eng. diagonal)'', which was ranked as the 12th most changed word according to the average rank criteria. The manual inspection showed that it is very often used in the football context of the ``diagonalna podaja (eng. diagonal pass)''. Since the word ``globinski (eng. deep)'', which was ranked as the fifth most changed word, was used frequently in the exact same football context, and made it to the final list of changed words, the word ``diagonalen'' was discarded.

To appease the evolutionary diversity criteria, the two native speakers were explicitly instructed to try to find words in different phases of usage change. This way, the created test set would not only evaluate the tested models' ability to derive a binary (i.e. changed/unchanged) prediction, but rather to distinguish between different stages and types of semantic changes. To achieve this, two native speakers were explicitly instructed to maximize the evolutionary diversity criteria by picking out words that exhibit varying degrees of semantic change, e.g., words with a clearly distinct main usage in two time periods, words that obtained one or more common, but not majority new usages, words that lost a significant, but not a majority new usage, and words that obtained or lost rare minority usages. While they were allowed to use the average rank criteria as a guidance, they were also explicitly instructed to rely as much as possible on their background knowledge of Slovenian language and manual context checking for the final word selection. The final manual selection included 49 words with diverse average ranking, i.e., it contained 8 out of 10 most changed words, 17 out of 50 most changed words, 26 out of 100 most changed words, 40 out of 500 most changed, and 49 out of 1,000 most changed words words. 

Additionally, we asked the two native speakers to check words with a major disparity in ranking by different systems for a possible inclusion in the final word list. These were words that ranked below the 1,000 most changed words threshold according to the average rank, but were nevertheless ranked very high by one of the systems. An example of this type of word is ``evro (euro)'', with an average rank of 5,245.7. This word was ranked as 8th by the clustering system proposed by \cite{montariol2021scalable}, as 14,849th by the {\em Nearest Neighbors} method, and as 880th by the \textit{SGNS+OP+CD} method. We figured that inclusion of words with such a disparity in the ranking would additionally increase the difficulty of the final test set and also allow for a more fine-grain comparison of different types of models, by revealing what kind of shifts they can or cannot detect. The manual inspection pointed out seven interesting words with a large disparity in the ranking, namely \textit{evro, kontaktirati, kotacija, poceniti, priklju\v{c}ek, zapreka, zmerno}, which were added to the list.

After that, 49 filler words (i.e., words with an average rank less than 1,000), were sampled from the dataset in a way so that they match the part of speech and frequency distribution of 49 identified changed words across both time periods, same as in \cite{kutuzov2021rushifteval}. More specifically, for each of the 49 manually selected words with identified usage change, we obtained the most similar counterpart with identical part of speech\footnote{CLASSLA-Stanza \citep{tervcon2023classla} part of speech tagger was used to obtain part of speech tags.} and of comparable (near identical) frequency in each time slice. By doing this, we ensured that part of speech and frequency information cannot be used to distinguish the target words from fillers. While the initial selection of filler words was done automatically, by randomly choosing one appropriate filler word for each changed word, the retrieved list of random filler words was manually checked for corpus artefacts by two native speakers, who were also instructed to check whether the number of topics covered by the main senses of the selected filler words were sufficiently diverse. The potentially problematic words were replaced with other appropriate randomly selected words until we obtained the final list of 49 filler words. 

The final word list contained 105 words, 49 changed words, 49 filler words and 7 words with a major disparity in ranking, all of which were single-word tokens (i.e. there were no multiword expressions).

\subsection{Annotation} 

For each target word in the list, we extracted 30 usage examples (sentences) from the dataset (corpus) from the years 1990--1997 and 30 usage examples from the year 2018. The extraction was performed by matching the lemma of each target word with the lemmatized forms in the corpus, ensuring that the inflectional variants were included. The sentences from both time periods were randomly matched (i.e., each pair contains a random sentence from 1990--1997 and a random sentence from 2018, both containing the same target word), resulting in 3,150 sentence pairs. 

The annotation was carried out by three linguistics students, selected through an internal call based on academic performance and prior annotation experience. They were not involved in the earlier stages of the project and were paid for their work. Each annotator independently annotated all 3,150 sentence pairs using a shared spreadsheet, where sentence pairs for the same word appeared consecutively, while the order of words was randomized. Each row contained the target lemma, a usage example from 1990--1997 and one from 2018, a field for the relatedness score, and a link to the relevant entry in the Slovenian reference dictionary \citep[SSKJ,][]{sskj} for help. No formal training was provided beyond written guidelines and a group discussion after a pilot round.  A sample of the annotated dataset is provided in Appendix~\ref{tbl:sample_data}.

The annotations followed the DURel framework \citep{schlechtweg2018diachronic}, which uses a scale from 1 to 4 to indicate the degree of semantic relatedness between usages:
\begin{itemize}
 \item 1 means that the uses in the sentences are not related to each other
 \item 2 means that the uses in the sentences are distantly related
 \item 3 means that the uses in the sentences are closely related to each other
 \item 4 means that both uses are identical (they have the same meaning)
\end{itemize}

The corresponding annotation guidelines (in Slovene) consisted of a short document in which the four categories were explained in more detail and illustrated with some prototypical examples. In particular, annotators were instructed to assign the label \textit{1 -- unrelated meaning} to cases where the two meanings of a word are completely unrelated, e.g. \textit{burka} as a dramatic composition (‘a farce’) on the one hand and \textit{burka} as a garment (‘a burqa’) on the other, whereby the equivalence in form is purely coincidental (homonymy). Label \textit{2 -- distantly related meaning} was assigned to sentence pairs with two different but cognitively, diachronically or otherwise seemingly related meanings of a word, e.g. \textit{dopisnica} as postcard on the one hand and \textit{dopisnica} as correspondent (reporter on site) on the other, both of which are derived from the concept of correspondence (communication through exchanging letters). In contrast, label \textit{3 -- closely related meaning} was assigned to pairs of phrases with very similar but not identical meanings, such as the adjective \textit{globinski} ‘deep’, which denotes something that goes far into the depths (e.g. a deep sea) on the one hand, or something with a strong effect (e.g. a deep cleaning) on the other, the difference in meaning being much more subtle and context dependent compared to sentence pairs labeled with 1 and 2. Finally, annotators were instructed to use the label 4 -- identical meaning - for sentence pairs where there is no significant difference in the meaning and syntactic context of a word, such as when using the noun \textit{razbitina} ‘wreck’ to describe a severely damaged vehicle, regardless of the vehicle type (e.g., car/train wreck).

After the preliminary round of annotations, the guidelines were expanded to provide additional clarifications regarding three specific groups of words that appear in the original list: named entities, such as \textit{Zenit} (a soccer club), names of team members, such as \textit{teli\v cki} ‘(Dallas) Mavericks’, and adjectives with homographic lemmas, i.e. lemmas that have the same spelling but a different pronunciation, such as \textit{testen}, which can refer to either a test (with \textit{testnega} as the genitive form) or a dough (with \textit{testenega} as the genitive form). While the annotators were instructed to discard the first group (named entities) by using the 0-label described below, the names of team members, which are usually lowercase in Slovene, were left in the dataset as typical examples of metaphorical derivations of meaning. Likewise, pairs of homographic adjectives were retained as examples of (loose) homonymy, especially since some are also diachronically related, such as \textit{vezen} ‘linking’ as in \textit{vezni \v clen} ‘linking element’ and \textit{vezen} ‘embroidered’ as in \textit{vezeni prt} ‘embroidered tablecloth’, both of which refer to the term \textit{vez} ‘a bond’.

The additional label \textit{0 -- not applicable} was included to mark instances that, in the opinion of the annotators, could not be labeled using the above scale (e.g. due to lack of context, ambiguity, preprocessing noise). However, this label was only used sparingly (3.2\% of all decisions), mainly examples of the word \textit{zenit}, where the mentions of the soccer club \textit{Zenit} were incorrectly matched to examples of the common noun \textit{zenit} ‘zenith’ in the preprocessing phase. Therefore, the word \textit{zenit} was removed from the dataset altogether, lowering the number of final sentence pairs from 3,150 to 3,120. Other 0-labeled judgments were retained in the dataset but excluded from the agreement calculations presented in the following paragraph.

\subsection{Inter-Annotator Agreement} 

In absolute terms, the three annotators completely agreed on 1,939 out of 3,120 decisions (62\%) with a Krippendorff's alpha (ordinal) of 0.721, somewhat higher than the inter-annotator agreement reported in related annotation campaigns focusing on this particular type of semantic disambiguation \citep{schlechtweg2018diachronic,kutuzov-pivovarova-2021-three,schlechtweg2020semeval,rodina-kutuzov-2020-rusemshift,d-zamora-reina-etal-2022-black}. Despite the inherently subjective nature of the task, the overall agreement was substantial: pairwise weighted Cohen’s $\kappa$ scores for labels 1--4 ranged from 0.63 to 0.70 and Spearman $\rho$ correlation scores ranged from 0.73 to 0.75 (see Table~\ref{tab:kappa}).

As expected, given the challenges of delineating word meaning \citep{hanks2013lexical}, annotators most often disagreed on the boundary between similar meanings --- specifically, whether two usages reflected a slight difference (label 3) or no difference at all (label 4), and whether the meanings were closely (label 3) or only distantly related (label 2). These two cases accounted for 55\% and 21\% of all disagreements, respectively (see Table~\ref{tab:confusion_matrix}), underscoring the graded nature of semantic relatedness. At the same time, the exceptionally high number of agreements on label 4 indicates a strong consensus in identifying cases of identical meaning, suggesting that annotators had a stable and shared understanding of clear semantic equivalence.

\begin{table}[h!]
\centering
\begin{tabular}{lccc}
\toprule
\textbf{Annotator pair} & \textbf{Cohen's $\kappa$} & \textbf{Spearman's $\rho$} \\
\midrule
\textbf{A1} vs. \textbf{A2} & 0.626 & 0.727\\
\textbf{A1} vs. \textbf{A3} & 0.698 & 0.746\\
\textbf{A2} vs. \textbf{A3} & 0.661 & 0.736\\
\bottomrule
\end{tabular}
\caption{Pairwise inter-annotator agreement on our dataset measured with Cohen's $\kappa$ and Spearman's rank correlation ($\rho$). Different annotators are labeled with A1 through A3.}
\label{tab:kappa}
\end{table}

\begin{table}[h!]
\centering
\begin{tabular}{lcccc}
\toprule
\textbf{A $\backslash$ B} & \textbf{1} & \textbf{2} & \textbf{3} & \textbf{4} \\
\midrule
\textbf{1} & \cellcolor{gray!20}114 & 1   & 0   & 0   \\
\textbf{2} & 6   & \cellcolor{gray!20}1,651 & 226 & 118 \\
\textbf{3} & 30  & 433 & \cellcolor{gray!20}448 & 609 \\
\textbf{4} & 5   & 167 & 866 & \cellcolor{gray!20}4,620 \\
\bottomrule
\end{tabular}
\caption{Confusion matrix summarizing pairwise annotation disagreements for labels 1--4 across all annotator pairs. Diagonal cells (in gray) indicate agreement; off-diagonal cells show disagreement.}
\label{tab:confusion_matrix}
\end{table}

\subsection{Semantic Change Scores and Observations} 

Finally, the annotated dataset was used to calculate the degree of semantic change for each word between two time periods using the COMPARE metric \citep{schlechtweg-etal-2019-wind}, which is a simple average of the relatedness scores for all sentence pairs and all three annotators. Essentially, higher scores are given to words with relatively stable meanings (e.g. the full score of 4 for words such as \textit{dokumentarec} ‘documentary’, \textit{metafora} ‘metaphor’ and \textit{odstavljen} ‘removed’), while words with changing or competing meanings receive lower scores (e.g. 1.2 for \textit{burka} ‘burqa/farce’, 2.1 for \textit{portal} ‘doorway/website’, 2.2 for \textit{replika} ‘replica/reply’). %

As expected, the resulting list shows a moderate correlation with the three baseline systems used to identify the target words (with Spearman’s Rho between 0.47 and about 0.53; see results in Section \ref{sec:evaluation} for details), but some discrepancies in the ranking of certain words can also be observed, such as the noun \textit{izkrcanje} `disembarkation', which was identified as a meaning-changing polysemous word by all three systems used in the creation of the original list, but received a score of 4.0 in the final gold standard list, meaning that for all 30 sentence pairs examined, all 3 annotators agreed that there was no difference in meaning. Future work is needed to investigate whether this is due to the limitations of the original word list selection on the one hand, or sentence selection and matching on the other. Nonetheless, the number of such discrepancies remains relatively small, especially when zooming in on the word list with the lowest score and the shift in meaning.

In particular, words showing the lowest score values align with previous research findings in Slovene lexical semantics \citep{gantar2018leksikalne,fiser2019distributional}, that recent semantic changes have mostly occurred in the context of new technologies with words like \textit{portal} (`website'), \textit{\v{c}arovnik} (`wizard'), \textit{ikona} (`icon'), and \textit{zakro\v{z}iti} (`to go viral'), and daily events, such as trending sports events \textit{vrag} (`devil'), \textit{teli\v{c}ek} (`maverick'), and \textit{plezalka} (`climber').
However, a more systematic analysis would be needed to determine whether the observed differences in word usage are primarily due to new meanings emerging or becoming more prominent in the language in general or are mainly influenced by data biases, such as variations in text sources across studied time periods.

\section{Semantic Change Detection Through Optimal Transport}
\label{sec:methodology}

In this section, we take a detailed look at the APD metric and present a simplified formulation of the metric. We show that, contrary to the previous claims, the metric is based on the vector aggregation strategy. A limitation of the APD metric is its tendency to detect spurious semantic changes in scenarios where no such changes are present. To mitigate this issue, we propose using optimal transport with the entropic regularization. We show how the proposed metric combines contributions from both the optimal transport and the APD to derive a more robust and accurate metric.

\subsection{Average Pairwise Distance Metric}
\label{sec:apd}
 
The Average pairwise distance (APD) metric was first introduced to quantify the density of a group of vectors \citep{sagi-etal-2009-semantic}. This density measure was used to identify polysemous words, as these words often appear in highly dissimilar contexts, resulting in low density and high APD scores.  
In subsequent work, APD was reformulated as a measure of semantic change in \cite{giulianelli-etal-2020-analysing} to address challenges related to the aggregation of all senses of a word into a single representation. 
The effectiveness of this metric is demonstrated through its robust performance across multiple semantic change detection datasets \citep{giulianelli-etal-2020-analysing,kutuzov2020uio,Wang2020UniversityOP,rachinskiy-arefyev-2022-black}. The formal method for calculating this metric is presented in Eq.~\ref{eq:apd-old}. Prior work \citep{giulianelli-etal-2020-analysing,periti-24-survey} has described the APD as a metric that, unlike others, does not rely on clustering or aggregation techniques; APD avoids these steps by directly computing the average distance between individual word usages. 
However, we show that this characterization is inaccurate. In Eq.~\ref{eq:apd-new}, we provide a simplified formalization of APD and derive an alternative form that clearly reveals its dependence on an implicit aggregation step. Although variants of APD using Euclidean and Canberra distance are proposed in \cite{giulianelli-etal-2020-analysing}, the most common variant is the one using Cosine distance ($d_{cos}$) and shown in Eq.~\ref{eq:apd-old}. It is defined as a measure of change of a target word between two time periods ($T_1$ and $T_2$) containing vector representations $\vec{p}$ from time period $T_1$ and $\vec{q}$ from time period $T_2$. The final form in Eq.~\ref{eq:apd-new} is defined as the dot product between the normalized average vectors from $T_1$ and $T_2$, represented by ${\overline{\hat{p}}}$ and ${\overline{\hat{q}}}$.
The final form of Eq.~\ref{eq:apd-new} uses the average of normalized vectors due to the equivalence of the cosine similarity and a dot product of unit vectors, even though the original vectors ($\vec{p} \in T_1$, and $\vec{q} \in T_2$) need not be normalized.

\begin{align}
\begin{split} \label{eq:apd-old}
APD(T_1, T_2) &= \frac{1}{|T_1| \ |T_2|} \cdot \sum_{\vec{p} \in T_1}\sum_{\vec{q} \in T_2}{d_{cos}(\vec{p}, \vec{q})}
\end{split}
\\
\begin{split} \nonumber
\hat{v} &= \frac{\vec{v}}{||\vec{v}||} \quad \overline{\hat{v}} = \frac{1}{|T|}\sum_{\vec{v} \in T}{\hat{v}} \\
APD(T_1, T_2) &= \frac{1}{|T_1| \ |T_2|} \cdot \sum_{\vec{p} \in T_1}\sum_{\vec{q} \in T_2}{(1 - \hat{p} \cdot \hat{q})} \\
 &= \frac{1}{|T_1| \ |T_2|} \cdot (|T_1| |T_2| - \sum_{\vec{p} \in T_1}\sum_{\vec{q} \in T_2}{(\hat{p} \cdot \hat{q})}) \\
 &= 1 - \frac{1}{|T_1| \ |T_2|}\sum_{\vec{p} \in T_1}  \sum_{\vec{q} \in T_2}{(\hat{p} \cdot \hat{q})} \\
 &= 1 - \frac{1}{|T_1| \ |T_2|}\sum_{\vec{p} \in T_1}{\hat{p}} \cdot \sum_{\vec{q} \in T_2}{\hat{q}} \\
 &= 1 - \frac{\sum_{\vec{p} \in T_1}{\hat{p}}}{|T_1|} \cdot \frac{\sum_{\vec{q} \in T_2}{\hat{q}}}{|T_2|}
\end{split}\\
\begin{split} \label{eq:apd-new}
 &= 1 - {\overline{\hat{p}}} \cdot {\overline{\hat{q}}}
\end{split}
\end{align}

Several consequences can be observed from Eq.~\ref{eq:apd-new}. The derived form demonstrates that APD is fundamentally a measure based on the aggregated vectors of corpus slices $T_1$ and $T_2$. Compared to the formulation presented in Eq.~\ref{eq:apd-old}, the computational efficiency of the new APD formulation is significantly improved. Specifically, the new approach exhibits a linear scaling of computational resources (both memory and compute) with respect to the size of the vector set, whereas the old method required quadratic scaling. By using an online version of the averaging algorithm, memory requirements for APD can be further decreased. This improvement implies that the new formulation can effectively handle much larger datasets and Monte Carlo estimation strategies proposed in \cite{sagi-etal-2009-semantic} are no longer required. 

Similar steps can be applied to all functions derived from APD like word \textit{variation coefficient} from \cite{addMoreClusters2020} to show that it equals a difference of the average vector Euclidean norm from a unit norm:
\begin{align}
\begin{split} \nonumber
\overline{p} &= \frac{1}{|T|} \sum_{\vec{p} \in T}{\vec{p}} \\
var_w(T) &= \frac{1}{|T|} \sum_{\vec{p} \in T} d_{cos}(\vec{p}, \overline{p}) \\
var_w(T) &= APD(T, \{\overline{p}\}) = 1 - {\overline{\hat{p}}}  \cdot {\hat{p}} = 1 - ||{\overline{\hat{p}}}|| \cdot \hat{p} \cdot \hat{p} 
\end{split} \\
var_w(T) &= 1 - ||{\overline{\hat{p}}}||
\end{align}

\noindent Difference between token embedding diversities (DIV) from \cite{kutuzov2020distributional} can also be simplified to the difference of average vector norms:

\begin{align}
\begin{split} \nonumber
DIV(T_1, T_2) &= \left\vert \frac{1}{|T_1|}\sum_{\vec{p} \in T_1} d_{cos}(\vec{p}, \overline{p}) - \frac{1}{|T_2|}\sum_{\vec{q} \in T_2} d_{cos}(\vec{q}, \overline{p}) \right\vert \\
DIV(T_1, T_2) &= \left\vert var_w(T_1) - var_w(T_2)\right\vert
\end{split} \\
DIV(T_1, T_2) &= \left\vert ||\overline{\hat{p}}|| - ||\overline{\hat{q}}|| \right\vert
\end{align}

\noindent Measure APD-OLD/NEW used by \cite{laicher-etal-2021-explaining} to estimate the average degree of polysemy is:
\begin{align}
\nonumber
APD_{O/N}(T_1, T_2) &= \frac{APD(T_1, T_1) + APD(T_2, T_2)}{2} \\
APD_{O/N}(T_1, T_2) &= 1 - \frac{||\overline{\hat{p}}||^2 + ||\overline{\hat{q}}||^2}{2}
\end{align}

The APD distance metric can be described relative to the COS metric, a simple cosine distance of the mean vectors of $T_1$ and $T_2$. As visible in Eq.~\ref{eq:cos-to-apd}, while the COS metric accounts solely for the angular component between the mean vectors by normalizing the cosine similarity by their norms, the APD metric additionally incorporates the magnitude of these vectors into its score (see Eq.~\ref{eq:apd-new}). In this sense, COS can be viewed as a special case of APD where only the directional component of the mean vector is considered.

\begin{equation}
COS(T_1, T_2) = d_{cos}(\overline{p}, \overline{q}) = 1 - \frac{{\overline{p}} \cdot {\overline{q}}}{||{\overline{p}}|| \ ||{\overline{q}}||} \label{eq:cos-to-apd}
\end{equation}

One limitation of the APD metric is that it detects some degree of semantic change even when there is none. This behavior can be understood from Eq.~\ref{eq:apd-new}, where the aggregation vectors $\vec{\overline{p}}$ and $\vec{\overline{q}}$ are computed. Unless all constituent vectors used in the computation of the average are identical, these averaged unit vectors will not have unit norm by definition. Consequently, any dispersion among the individual vectors contributes to an elevated APD score, regardless of whether a genuine semantic change is present. This implies that the APD score tends to increase with greater dispersion in the vectors, regardless of any underlying semantic changes.

\subsection{Optimal Transport With Entropic Regularization}
\label{sec:ot-with-entropy}
In order to address this limitation, we propose to place weights on individual pairs of distances such that the closest matches between $T_1$ and $T_2$ are assigned a stronger weight. Local optimization of such matching can assign the strongest weight of multiple items from $T_1$ to a single item in $T_2$. This concentration of weights to nearest neighbours is undesirable as it would underestimate the total distance between $T_1$ and $T_2$. Our goal is to assign weights such that an item from $T_1$ is most strongly connected to a distinct pair from $T_2$, and the total distance between all pairs is minimized.

The above description aligns well with the optimal transport (OT) problem, a mathematical framework to compare probability distributions. Introduced with the work of \cite{monge1781memoire}, who sought to minimize the cost of transporting materials from one location to another, OT has evolved into a powerful tool with applications across a wide range of fields. The optimal transport addresses the problem of finding the most efficient way to transform one probability distribution into another, by minimizing a given cost function over all possible transportation plans.

A fundamental application of optimal transport lies in the computation of distances between probability distributions or histograms, commonly referred to as the Wasserstein Distance (WD) or Earth Movers Distance (EMD). As demonstrated by \cite{montariol2021scalable}, this framework can be used for semantic change detection. Specifically, the occurrences of a word are partitioned into a set of semantic clusters, with the relative sizes of these clusters representing the word's semantic probability distribution as a histogram. The evolution of this distribution from one time slice $T_1$ to another, $T_2$, serves as an indicator of semantic shift. In this context, the optimal transport problem is defined as the minimal cost required to transform the semantic distribution at time slice $T_1$ into that of $T_2$.

Solving an optimal transport problem allows us to find the best mapping between two groups of objects, or the best match between elements of a complete bipartite graph such that the source is completely mapped to the target. In this context, ``best'' implies a loss function that is minimized. In the case of the optimal transport problem, this loss is a linear sum of the losses incurred by each individual mapping between pairs across two sets. The solution to the optimal transport problem is a transport plan that assigns the source to the destination in such a way that the transported amount weighted by the transport costs is minimized.

We can formally define the transport plan ($A = \{a_{i,j}\}, \forall i,j$) that minimizes the distance between two empirical distributions represented by a set of vectors $\vec{p_i} \in T_1$ and $\vec{q_j} \in T_2$ as a minimum of a linear combination of a planned assignment ($a_{i,j}$) and a distance between individual vectors ($\vec{p_i}$, and $\vec{q_i}$). As optimal transport is a measure between probability distributions, we will normalize the transport plan such that the sum across all rows, as well as a sum across all columns, equals \textbf{1}.
\begin{align}
\begin{split} \label{eq:optim-transport}
d_{OT}(T_1, T_2) &= \min\limits_A {\sum_{p_i \in T_1}\sum_{q_j \in T_2} a_{i,j} \cdot d(p_i, q_j)}\\
& \qquad \text{such that} \\
\sum_{i}a_{i,j} &= 1, \ \forall j \quad \text{and} \quad \sum_{j}a_{i,j} = 1, \ \forall i
\end{split}
\end{align}

Without normalizing the transport plan in the equation above, the resulting transport cost becomes sensitive to the number of vectors in $T_1$ and $T_2$. This sensitivity renders it unsuitable for comparing words with unequal occurrence counts.

The optimal transport is a powerful framework for comparing probability distributions, but it can be computationally challenging and sensitive to noise. Entropic regularization of optimal transport \citep{NEURIPS2019_159c1ffe} addresses these limitations by incorporating an entropy term into the standard optimal transport problem. This transformation converts the original linear program into a more computationally tractable form, enabling efficient algorithms such as those based on Sinkhorn iterations. Furthermore, the entropy term acts as a regularizer, reducing the impact of noise on the transport plan \citep{RIGOLLET20181228}. The general formulation for entropic-regularized optimal transport cost between two empirical distributions $T_1$ and $T_2$ is provided in Eq.~\ref{eq:optim-transport-reg}.

\begin{align}
\begin{split} \label{eq:optim-transport-reg}
d_{OT_\lambda}(T_1, T_2) &= {d_{OT}}(T_1, T_2) - \lambda \cdot h(A)\\
& \text{where}\\
h(A) &= -\sum_{i,j}{a_{i,j}\cdot log(a_{i,j})}
\end{split}
\end{align}

The entropy-regularized optimal transport problem introduces a regularization term that encourages the resulting transport plan to favor transport plan with higher entropy. The regularization strength is controlled by the parameter $\lambda$, where larger values of $\lambda$ increase the entropy of the resulting plan. The transport plan achieving the maximum entropy corresponds to uniform weights across all pairs between $T_1$ and $T_2$, which aligns precisely with the formulation of the APD metric (cf., Eq.~\ref{eq:apd-old}).

\begin{figure}[htb]
\centering
    \subfloat{\includegraphics[width=0.35\textwidth]{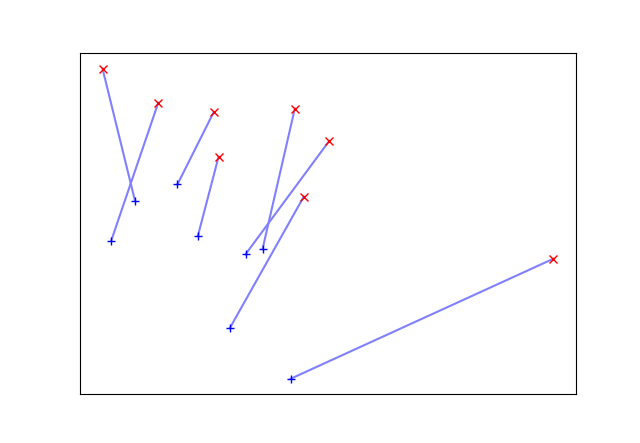}\hspace{-0.5cm}}
    \subfloat{\includegraphics[width=0.35\textwidth]{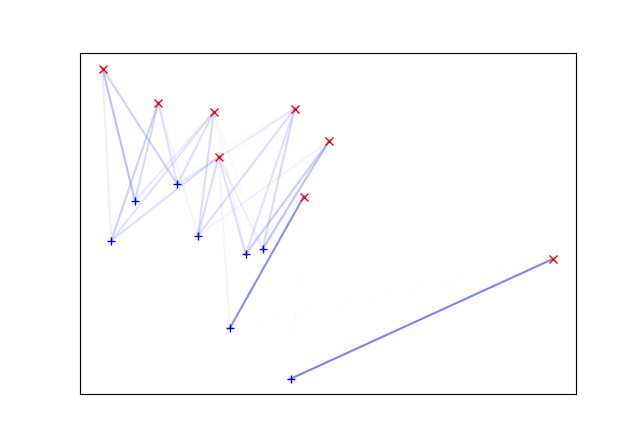}\hspace{-0.5cm}}
    \subfloat{\includegraphics[width=0.35\textwidth]{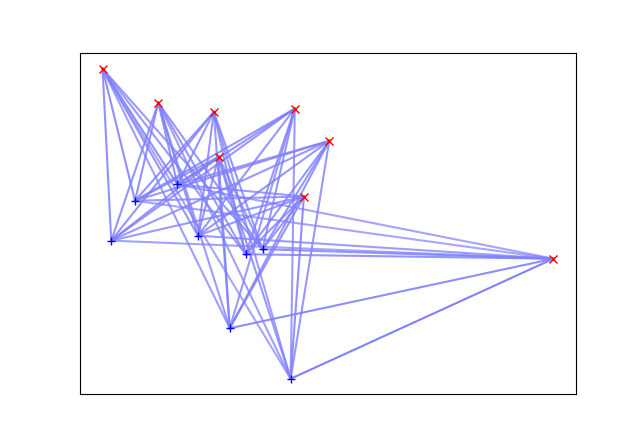}}
\caption{Effect of the entropic regularization on the result of optimal transport results. First subfigure shows optimal transport results without regularization, the one in the middle uses small regularization ($\lambda=0.7$) and the last one uses very high value ($\lambda=100$) showing near-equivalence with APD.}
\label{fig:regularizations}
\end{figure}

The regularized transport plan can be interpreted as a combination of two extremes: one where each item in $T_1$ is paired with exactly one item in $T_2$, and another where every item in $T_1$ is assigned to every item in $T_2$. The parameter $\lambda$ serves as the balancing factor between these two scenarios. We illustrate this effect in Figure~\ref{fig:regularizations} where two sets of points are shown in blue and red color, respectively. First subfigure depicts results of optimal transport problem without regularization, red and blue samples are connected in pairs such that a sum of connection lengths is minimal. Adding a small amount of regularization, shown in the second subfigure, changes the transport plan such that each item from one set can be connected to multiple items from another set. Stronger weight of connections is assigned between pairs that are closer. The last subfigure shows optimal transport problem with a very large amount of regularization where this regularization takes over the entire optimization of a transportation problem and produces a transport plan that almost matches the APD.\footnote{In order to reach exact values produced by APD, we would have to use much higher values of regularization. In theory it would take an infinite amount of regularization to completely overtake the finite amount of transport plan from the distances between samples, although this is reached much sooner due to numerical inaccuracies.}

The central part of modeling the problem as a transport problem is the determination of a cost matrix. The cost matrix represents the unit cost of transport or a distance between each point of the source and the destination. In our application of optimal transport to semantic change detection, both the source and the destination are a set of numerical vectors obtained from the deep language model. A common approach to compare such vectors is to compute a cosine distance \citep{Reimers2019SentenceBERTSE, addMoreClusters2020, montariol2021scalable} between them. Cosine distance has two issues: it is not a proper metric as it does not satisfy the triangle inequality,\footnote{It can, however, be converted to one by the inverse cosine ($arccos$) function turning it to distance metric proportional to the angular distance.} and it is not a convex function.\footnote{Which can be important in some optimization procedures like OT to guarantee a unique solution to the problem.} Nevertheless, it has empirically been shown to perform well in comparison of high dimensional vectors.

\subsubsection{Detecting Semantic Change through Regularized Optimal Transport}

Our approach leverages regularized optimal transport to rank words based on their semantic shift between two corpora ($T_1$ and $T_2$) and relies on representations from BERT-like models. Such models contain multiple hidden layers on which word representations can be based on. Typically, final or near-final layers are preferred, as probing tasks demonstrate that earlier layers are less suitable for capturing semantic information \citep{Zhang2020BERTScoreET, liu-etal-2019-linguistic}.

The process begins with tokenization, where text is represented as a sequence of model-specific tokens. A single word will be represented either with a single token if this word is part of the model's vocabulary, or with a sequence of subword tokens. In order to represent the whole word, regardless of the number of tokens it requires, common way is to average individual token representations belonging to this word. 

Consider a BERT-like model with $K$ hidden layers and an input sequence containing $N$ tokens (including special tokens for input text boundaries). Each token has $(K+1)$ representations: one from the initial embedding layer ($L_0$) and $K$ from the subsequent hidden layers. For a word represented using $n$ subword tokens, its representation at layer $k$ is computed as:

\begin{equation}
  \vec{w}_k = \frac{1}{n} \sum_{i=j}^{j+n-1} \vec{h}_{i,k},
\end{equation}
where $\vec{h}_{i,k}$ is the representation of the $i$-th token at layer $k$, and $j$ is the starting index of the word's tokens.
For representations based on multiple hidden layers, we additionally average across a range of layers ($k_1$ to $k_2$):

\begin{equation}
  \vec{w}_{k_1-k_2} = \frac{1}{(k_2 - k_1 + 1)} \sum_{k=k_1}^{k_2} \vec{w}_k.
\end{equation}

To rank words by their semantic shift between $T_1$ and $T_2$, we order them by the optimal transport cost. First, for each word we find all of their occurrences in  $T_1$ and $T_2$. Next, we infer their contextual representations with regard to the chosen layers. Finally, for each word we compute the cost of regularized optimal transport between $T_1$ and $T_2$ using some predefined regularization factor ($\lambda$) using Eq.~\ref{eq:optim-transport-reg}. The words are ordered by the resulting cost, with higher values indicating greater semantic shift.

\section{Experimental Setup}
\label{sec:experimental_setup}
To analyze our approach, we evaluate it using the newly created dataset described in Section~\ref{sec:dataset}. We infer contextual representations of target words using the 12-layer Slovene SloBERTa model \citep{ulvcar2021sloberta} on sentences from the Gigafida 2.0 corpus (see Section~\ref{sec:corpus-selection}).

Our proposed optimal transport-based approach is compared with several baseline methods described in Section~\ref{sec:related_work}. We evaluate two baseline methods based on static embeddings, \emph{SGNS+OP+CD} method \citep{hamilton2016cultural} and \emph{Nearest Neighbors} method proposed by \cite{gonen-etal-2020-simple}. The \emph{SGNS+OP+CD} method combines skip-gram negative sampling (SGNS), orthogonal Procrustes alignment (OP), and cosine distance (CD) to measure semantic change, and the \emph{Nearest Neighbors} method measures semantic change for a word as the size of the intersection between its nearest neighbor lists in two time periods. 

For clustering-based baselines, we follow related work \citep{montariol2021scalable,giulianelli-etal-2020-analysing} and identify k=5 clusters using k-means. The final word representation is derived as the average of the last four layers ($L_{9-12}$) of the SloBERTa model. Semantic change is then measured using two metrics: Jensen-Shannon divergence (JSD) between cluster assignments, and Wasserstein distance (WD) between cluster centroids.

In addition to clustering-based methods, we evaluate aggregation-based approaches by comparing the average word representations in $T_1$ and $T_2$. Individual word representations are taken from the final layer ($L_{12}$) of the model, similar to the related work \citep{Arefyev2021DeepMistakeWS,pomsl-lyapin-2020-circe,10.1007/978-3-030-72610-2_13,rachinskiy-arefyev-2022-black}.
Such average representations are compared using APD and PRT metrics. We include variants of the APD metric where only the direction ($APD_D$) or magnitude ($APD_M$) of the aggregate vector is considered. The $APD_D$ variant, which considers only direction, is equivalent to the cosine distance between aggregate vectors (see Equations \ref{eq:apd-new} and \ref{eq:cos-to-apd}). In Section~\ref{sec:apd}, we have discussed one limitation of APD, that it detects semantic change even when there is none. To provide more insight to this behavior, we introduce an additional variant that is not supposed to measure any change ($APD_C$), defined as:
\[
APD_C(T_1, T_2) = APD(T_1 \cup T_2, T_1 \cup T_2)
\]
where $T_1$ and $T_2$ represent the sets of word vectors in two time periods, and $T_1 \cup T_2$ represents their union. We can also note a similarity of $APD_C$ to the APD-OLD/NEW metric \citep{laicher-etal-2021-explaining} that expresses an average between $APD(T_1, T_1)$ and $APD(T_2, T_2)$.

Finally, we evaluate our approach using the cost of regularized optimal transport as described in Section~\ref{sec:methodology}. To compute the semantic change between $T_1$ and $T_2$, we first calculate the cosine distance between individual vector pairs in $T_1$ and $T_2$. We evaluate a range of regularization values ($\lambda$) using the word representations from the final layer ($L_{12}$) of the model. We leverage the PythonOT library \citep{flamary2021pot} to compute the optimal transport map and associated cost.

While in Table~\ref{tbl:results} we report the results using embeddings based on $L_{12}$ due to common use of the final layer to represent semantic information, in our preliminary experiments we observed that neither $L_{12}$ nor the average of the last four layers ($L_{9-12}$) yielded optimal results.
Instead, using a layer just before the final one ($L_{11}$) produced better performance.
To validate this observation, we evaluate selected metrics across all layers (see Section \ref{sec:evaluation}) independently and in standard combinations.
Some related work, like \cite{montariol2021scalable,kashleva-etal-2022-black,addMoreClusters2020}, uses finetuning on the masked language modeling (MLM) task to adapt the pretrained model to the target text distribution. Others, like in \cite{laicher-etal-2021-explaining,periti-etal-2022-done}, do not perform this step. Since our target dataset (selected from the Gigafida 2.0 corpus, as described in Section~\ref{sec:dataset}) is already part of the SloBERTa pretraining data \citep{ulvcar2021sloberta}, domain adaptation is not necessary, so we do not perform any finetuning.

We evaluate all systems using Spearman's rank correlation ($\rho$) with the gold standard rankings obtained via the COMPARE metric, as described in Section~\ref{sec:dataset}.

\section{Results}
\label{sec:evaluation}

\begin{table*}[!t]
\centering
\caption{Spearman's rank correlation between system output and ground truth rankings.}
\resizebox{0.6\textwidth}{!}{
\begin{tabular}{lc|c}
    \hline
    \multicolumn{2}{l|}{\bf Approach} & {\bf Spearman's rank correlation ($\rho$)}\\
    \hline\hline
    \multicolumn{3}{l}{\textsc{Static embedding methods}}  \\\hline
    \multicolumn{2}{l|}{\bf SGNS+OP+CD} & 0.477 \\
    \multicolumn{2}{l|}{\bf Nearest Neighbors} & 0.527 \\
    
    \hline\hline
    \multicolumn{3}{l}{\textsc{Clustering-based methods}}  \\\hline
    {\bf JSD} & $L_{9-12}$ & 0.458\\
    {\bf WD} & $L_{9-12}$ & 0.523\\
    
    \hline\hline
    \multicolumn{3}{l}{\textsc{Aggregation-based methods}}  \\\hline
    {\bf APD} & ${L_{12}}$ & 0.594\\
    {\bf APD$_D$} & $L_{12}$ & 0.529\\
    {\bf APD$_M$} & $L_{12}$ & 0.328\\
    {\bf APD$_C$} & $L_{12}$ & 0.406\\
    {\bf PRT} & $L_{12}$ & 0.527\\
    
    \hline\hline
    \multicolumn{3}{l}{\textsc{Optimal transport based (ours)}}  \\\hline
    {\bf OT$_{\lambda=0.0}$} & $L_{12}$ & 0.602 \\
    {\bf OT$_{\lambda=0.1}$} & $L_{12}$ & \bf{0.648} \\
    {\bf OT$_{\lambda=0.2}$} & $L_{12}$ & {0.632} \\
    {\bf OT$_{\lambda=0.3}$} & $L_{12}$ & {0.622} \\

    \hline\hline
\end{tabular}
}
\label{tbl:results}
\end{table*}
We present the results of our evaluation on the dataset described in Table~\ref{tbl:results} of Section~\ref{sec:dataset}. The table displays Spearman's rank correlation coefficients ($\rho$) between various system outputs and ground truth rankings. This correlation measure indicates how closely the systems' rankings align with the expected results, with higher values reflecting better agreement.
Out of baseline methods, APD achieves the highest score (0.594), which was expected given good performance on various shared tasks in related work.
Clustering based methods achieved score (0.523) comparable to the one achieved by the \emph{Nearest Neighbors} method based on static embeddings (0.527). We expected higher score from the clustering-based methods because they leverage contextual vectors of large pretrained language model, in contrast to the much simpler static embedding baseline.
Although finetuning should not be needed with this data (see Section~\ref{sec:experimental_setup}), we additionally evaluate clustering based methods on a model finetuned for ten epochs on the masked language modeling (MLM) task.
Empirically, we notice a small and inconsistent effect ($|\Delta \rho | < 0.02$), achieving a JSD score of 0.477 (up from 0.458), and a WD score of 0.504 (down from 0.523). Overall, the results are consistent with the results reported in the related work \citep{martinc2021embeddia}.

The proposed method based on optimal transport, which is described in Section \ref{sec:methodology}, achieves the highest score on this dataset compared to the best baseline (APD) approach.
Although the optimal transport without any regularization (OT$_{\lambda=0.0}$) slightly outperforms the baseline APD methods (0.602 compared to 0.594), it is clearly beneficial to use some regularization, as the highest result (0.648, up from 0.602) is achieved with $\lambda=0.1$. Although we cannot point to the exact reason for this increase, work of \cite{RIGOLLET20181228} shows that this type of regularization reduces the impact of the noise on the transport plan, which might explain this result. Further increase of the regularization slightly decreases the score, and with a large enough values of $\lambda$ it would converge to the baseline score of APD (see Section~\ref{sec:ot-with-entropy}).

As we noted in the previous section, our preliminary results indicated that the final layer ($L_{12}$) is not an optimal choice. For this reason, we evaluate chosen measures across all layers and their standard combinations and present those results in Table~\ref{tab:results_across_layers}.

\begin{table}[htb]
\centering
\caption{Spearman's rank correlation ($\rho$) for different evaluation metrics across model layers and layer combinations. Generally best performing layer ($L_{11}$) is highlighted with exceptions marked ($^*$).}
\label{tab:results_across_layers}
\begin{tabular}{l|lllllllll}
Layer & APD & $APD_{D}$ & $APD_M$ & $APD_C$ & PRT & $OT_{\lambda=0}$ & $OT_{{\lambda=0.1}}$ & $OT_{\lambda=0.2}$ & $OT_{\lambda=0.3}$ \\
\hline
$L_0$ & 0.257 & 0.357 & 0.192 & 0.174 & 0.348 & 0.328 & 0.351 & 0.357 & 0.345 \\
$L_1$ & 0.337 & 0.425 & 0.229 & 0.234 & 0.422 & 0.441 & 0.517 & 0.465 & 0.415 \\
$L_2$ & 0.377 & 0.433 & 0.235 & 0.250 & 0.428 & 0.469 & 0.514 & 0.472 & 0.449 \\
$L_3$ & 0.474 & 0.444 & 0.304 & 0.336 & 0.444 & 0.516 & 0.588 & 0.544 & 0.506 \\
$L_4$ & 0.567 & 0.473 & 0.283 & 0.368 & 0.472 & 0.535 & 0.606 & 0.592 & 0.586 \\
$L_5$ & 0.612 & 0.491 & 0.272 & 0.396 & 0.495 & 0.534 & 0.607 & 0.611 & 0.611 \\
$L_6$ & 0.583 & 0.490 & 0.184 & 0.319 & 0.489 & 0.524 & 0.585 & 0.588 & 0.587 \\
$L_7$ & 0.627 & 0.512 & 0.237 & 0.385 & 0.510 & 0.544 & 0.596 & 0.616 & 0.617 \\
$L_8$ & 0.646 & 0.525 & 0.257 & 0.395 & 0.529 & 0.565 & 0.619 & 0.637 & 0.640 \\
$L_9$ & 0.656 & 0.551 & 0.303 & 0.411 & 0.553 & 0.592 & 0.636 & 0.643 & 0.649 \\
$L_{10}$ & 0.694 & 0.560 & 0.416 & 0.498 & 0.560 & 0.630 & 0.669 & 0.687 & 0.687 \\
\cline{2-10}
$L_{11}$ & 0.705 & 0.565 & 0.446 & 0.532 & 0.567 & 0.651 & 0.691 & 0.706 & {0.707} \\
\cline{2-10}
$L_{12}$ & 0.594 & 0.529 & 0.328 & 0.406 & 0.527 & 0.602 & 0.648 & 0.632 & 0.622 \\
$L_{9-11}$ & 0.693 & 0.569$^*$ & 0.403 & 0.497 & 0.568$^*$ & 0.623 & 0.669 & 0.688 & 0.687 \\
$L_{10-12}$ & 0.699 & 0.569$^*$ & 0.418 & 0.518 & 0.569$^*$ & 0.632 & 0.679 & 0.695 & 0.697 \\
$L_{9-12}$ & 0.691 & 0.571$^*$ & 0.400 & 0.501 & 0.573$^*$ & 0.626 & 0.673 & 0.687 & 0.693 \\
\end{tabular}
\end{table}

According to Table~\ref{tab:results_across_layers}, in general, the best performing layer is $L_{11}$. There are some exceptions to this rule ($APD_D$ and PRT), but the difference is very small with $APD_D$ achieving 0.571 (up from 0.565) and PRT achieving 0.573 (up from 0.567) compared to $L_{11}$. All other methods benefit from representations chosen from $L_{11}$, with notable increases by $APD_C$ achieving 0.532 (up from 0.406), APD achieving 0.705 (up from 0.594), and $OT_{\lambda=0.3}$ achieving 0.707 (up from 0.622). Basing the results off the $L_{11}$, the difference between APD and our proposed method ($OT_{\lambda=0.3}$, in this case) is negligible (0.705 for APD and 0.707 for $OT_{\lambda=0.3}$). 
Such a small difference can also be explained by both of those results reaching a score almost on the level of human annotators, as median agreement between two annotators in Table~\ref{tab:kappa} achieves Spearman's $\rho=0.736$.

Although Table~\ref{tbl:results} (using $L_{12}$) demonstrates the best result with $\lambda=0.1$ and Table~\ref{tab:results_across_layers} (using $L_{11}$) shows peak performance with $\lambda=0.3$, our analysis reveals that the choice of the regularization parameter $\lambda$ is not overly sensitive. In Appendix~\ref{sec:more-results} we provide additional results for our method with the range of regularization values $\lambda$ from 0.05 to 0.95.
These results demonstrate that performance remains very close across the range of values from $\lambda=0.15$ to $\lambda=0.95$, with differences in Spearman's rank correlations across this range being $\Delta\rho\leq 0.01$.

\begin{figure}[htb]
\includegraphics[width=0.7\linewidth]{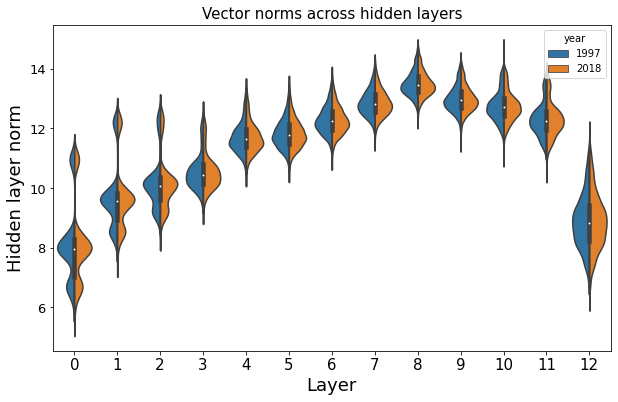}
\centering
\caption{Analysis of the magnitudes of hidden state representations across layers. For each hidden layer, we collect target word representations and plot their Euclidean norms.}
\label{fig:layer-norms}
\end{figure}

The observed decline in performance for $L_{12}$ and its clear departure from the otherwise consistent trend of increasing performance with higher layers is interesting. While prior work has explored various layer combinations, no systematic evaluation has specifically identified $L_{11}$ as an optimal choice or highlighted the notable decrease in performance for $L_{12}$.
From Table~\ref{tab:results_across_layers}, it is evident that the inclusion of $L_{12}$ in layer combinations has a minimal impact on improving the combined performance scores.
To gain more insights into the influence of the final hidden layer on semantic vectors, we analyze the magnitudes of these vectors across different model layers. As depicted in Figure~\ref{fig:layer-norms}, the Euclidean norms of word representations evolve as they propagate through the network. The norm starts at its minimum value at $L_{0}$, reaches a peak at $L_{8}$, experiences a gradual decline from $L_{9}$ to $L_{11}$, and undergoes a significant drop at $L_{12}$. Notably, $L_{11}$ has a 39\% higher average norm compared to the final layer ($L_{12}$).

Since the standard method for combining word representations from different layers involves averaging those vectors \citep[e.g.,][]{montariol2021scalable, martinc2021embeddia}, the final layer's contribution is diminished when averaged with layers that have larger norms.

\section{Conclusion and Future Work}
\label{sec:conclusion}

In this research, we have created a new dataset for training and evaluating semantic change detection systems for Slovene, a low-resource Slavic language with 2 million speakers. We use this dataset to evaluate several unsupervised semantic change detection models, including a novel model based on optimal transport that outperforms baseline models.

In the absence of previous work on semantic change in Slovene, which did not allow us to obtain a list of manually identified Slovene words with observed semantic change in a given time period, we decided to find these words semi-automatically by manually selecting words from a list of candidate words proposed by three systems for automatic change detection, similar to \cite{rodina-kutuzov-2020-rusemshift,schlechtweg2020semeval,kutuzov2021rushifteval}. For this reason, it is quite possible that the final target list is missing some words that are subject to a significant semantic change and were overlooked by all systems. While this does not affect the function of the constructed dataset as an evaluation dataset for semantic change detection systems, it does impose some limitation on the use of the dataset as a historical linguistic resource. For this reason, we plan to expand the target list in the future to include words identified by newly developed semantic change detection models as well as manually identified words to make it more complete.

During creation of a list of semantically changed words, some words were flagged as semantically changed by all three systems and later manually annotated as not having any semantic change. Such words with a large discrepancy between automated procedure and manual verification are of special interest as they have the ability to highlight the edge-cases of automated systems and provide insights toward their improvement. On the other hand, this can also be just a limitation of the original word list and an artefact of random sampling of sentence pairs. We leave a thorough analysis of a rationale behind this effect for future work.

In the SemEval-2020 Task 1: Unsupervised Lexical Semantic Change Detection shared task \citep{schlechtweg2020semeval}, a large discrepancy was found between the results of the different systems participating in the shared task for different languages. In particular, they pointed out the poorer performance of systems based on contextual embedding models compared to systems based on static embeddings. On the other hand, with the exception of APD, we observe a consistent performance of the different baseline systems on the new dataset and no major differences in the performance of systems based on static and contextual embeddings. Baseline approach using APD clearly outperforms other baseline approaches.
It has already been established that clustering of token representations does not lead to meaning-specific clusters \citep{addMoreClusters2020, NEURIPS2019_159c1ffe}. Moreover, as shown by \cite{addMoreClusters2020}, the task of detecting semantic change is very sensitive to the number of clusters, and the usual methods for optimization using the Silhouette score do not correlate with better performance.
Although APD was introduced in a paper that argues against vector averaging \citep{giulianelli-etal-2020-analysing}, and presents APD as a method that avoids this step, we show that this characterization is incorrect. Based on our analysis in Section~\ref{sec:apd}, we show that APD is implicitly defined as a dot product of average vector representations (see Eq.~\ref{eq:apd-new}).
Regardless of this characterization, performance of APD remains strong on our dataset.
In order to better understand the behavior of APD and not to propose other measures of semantic change, we introduce several APD variants.
As can be observed from Table~\ref{tab:results_across_layers}, APD significantly improves over other baselines and APD variants that take into account only the direction ($APD_D$, equivalent to the measure based on the cosine distance between average representations) or the magnitude ($APD_M$, equivalent to the measure based on vector norm of an average vector). This is an interesting observation as cosine distance (that ignores the norm of the vector) is regularly used to compare the representations from large language models.
The magnitude of the average vector reflects the dispersion of the individual vectors and including it clearly improves the performance on our task. Using just the magnitude component ($APD_M$) also reaches the non-trivial performance (0.446, from Table~\ref{tab:results_across_layers}). The variant that was evaluated with a union of vectors from $T_1$ and $T_2$, the setting without any change ($APD_C$), achieved even higher score (0.532, from Table~\ref{tab:results_across_layers}), comparable with the highest performance of clustering-based methods (0.523, from Table~\ref{tbl:results}) and methods based on static embeddings (0.527, from Table~\ref{tbl:results}). This result empirically confirms the observation that APD detects some change even when there is none. Although $APD_C$ performance is, as expected, much lower than approaches based on APD or OT, it is still unexpectedly high. Due to its simplicity -- just a vector norm of the average vector -- it might be suitable for fast filtering of word candidates that will be later evaluated using much slower and much more accurate methods. We leave this line of work for the future.

Our method to measure the semantic change based on the cost of the regularized optimal transport unexpectedly connects the APD metric with optimal transport through the entropic regularization of a transport plan. In Section~\ref{sec:ot-with-entropy}, we discuss how this approach addresses the identified limitation of APD. The strength of the proposed approach is empirically shown to outperform all other baseline methods, although sometimes with negligible differences, especially if the performance already reaches levels achieved by human annotators. Although regularization requires an additional parameter ($\lambda$), we have shown that the results are stable across a wide range of regularization values.

It has been shown \citep{Zhang2020BERTScoreET, liu-etal-2019-linguistic, turton-etal-2021-deriving} that the choice of a hidden layer for optimal performance on semantic tasks depends on the model family and the overall depth of the model, and in all cases the final layer was not the optimal choice. Our experiments provide additional evidence for this conclusion and performance of individual layers, as presented in Figure~\ref{fig:layer-norms}, match results published by \cite{turton-etal-2021-deriving}. The semantic change detection performance is significantly lower when the final hidden layer ($L_{12}$) is used to encode the semantic meaning of the word, compared to the previous layers. This observation calls into question the effectiveness of the usual approach of using the final layer embedding by itself, or as a part of the averaged representation, to obtain a semantic representation in the unsupervised semantic tasks \citep{Devlin2019BERTPO}. Nonetheless, our layer averaging experiments also show that including the final layer in the semantic representation does not significantly influence the performance of our approach. 
Our analysis suggests that this is due to the fact that the final layer of the SloBERTa model has a significantly smaller vector norm than the previous layers, which indicate that it contributes less to the final averaged representation. Future work should take a closer look at this phenomenon and determine if the mismatch of the final layer norm is a general trend among BERT-like models, a specifics of the SloBERTa model (a RoBERTa architecture pretrained on Slovene) or an artefact of the SloBERTa training procedure.
In future work, we plan to explore the use of generative Large language models (LLMs) for semantic change detection. These models could leverage their understanding of context to identify semantic shifts more accurately, particularly in low-resource languages like Slovene.

\section*{Data Availability}
The dataset is available via CLARIN.SI\footnote{Semantic change detection dataset for Slovenian 1.0, available from: \url{http://hdl.handle.net/11356/1651}} and the code via GitHub\footnote{The code used in experiments, available here: \url{https://github.com/sharpsy/slovene-OT-semchange}}.

\section*{Acknowledgements}
The authors acknowledge the financial support from the Slovenian Research and Innovation Agency for research core funding for the programmes Knowledge Technologies (No. P2--0103) and the Language Resources and Technologies for Slovene (No. P6--0411) as well as projects Large Language Models for Digital Humanistics (No. GC-0002), Embeddings--based techniques for Media Monitoring Applications (No. L2--50070), Computer-assisted multilingual news discourse analysis with contextual embeddings  (No.  J6--2581), and Hate speech in contemporary conceptualizations of nationalism, racism, gender and migration (No. J5--3102). We also acknowledge the project Development of Slovene in a Digital Environment co-financed by the Republic of Slovenia and the European Union under the European Regional Development Fund --- The project is being carried out under the Operational Programme for the Implementation of the EU Cohesion Policy in the period 2014--2020.

\clearpage
\section*{Appendix}

\subsection*{Sample of Annotated Data}
\label{tbl:sample_data}
\begin{table}[!ht]
\caption{A sample of the Slovene dataset for semantic change detection with a target word, year of the sentence and scores from three annotators. Below each sentence, an English translation is provided with a target word marked with an underline. Higher annotation scores correspond to a closer semantic match (less change) between two sentences.}
\begin{tabular}{l|cp{0.78\textwidth}c}
\textbf{Word} & \textbf{Year} & \textbf{Sentence} & \textbf{Scores} \\
\hline
\multirow{10}{3.8em}{globinski} & 1997 &
Pri plazovih sprijetega snega so to pogosto sti\v cne ploskve med starim-in-novim snegom ter skrajno labilne \v sibke plasti zasne\v zenega povr\v sinskega ali globinskega sre\v za \newline
(\textbf{eng}.) \textit{In the case of avalanches of stuck snow, these are often contact surfaces between old and new snow and extremely labile weak layers of snow-covered surface or \underline{deep} snow.} & \multirow{10}{3em}{2 3 2} \\
& 2018 & Veliko je izdelkov , s katerimi lahko nadomestimo draga globinska \v cistila za obraz . \newline
(\textbf{eng}.) \textit{There are many products that can be used to replace expensive \underline{deep} facial cleansers.} &  \\ \hline

\multirow{6}{3em}{burka} & 1997 & Burka je re\v c , ki jo lahko najla\v ze uprizorimo , in sicer iz dveh razlogov . \newline
(\textbf{eng.}) \textit{\underline{Burlesque} is the easiest thing to stage, for two reasons.}& \multirow{6}{3em}{1 1 1}\\
& 2018 & V primeru potrditve referenduma , bo Gallen drugi \v svicarski kanton , v katerem bodo prepovedali burke in nikabe , pred dvema letoma so to naredili v Ticinu . \newline
(\textbf{eng}.) \textit{If the referendum is approved, Gallen will be the second Swiss canton in which \underline{burkas} and niqabs will be banned, two years ago they did so in Ticino.} &  \\ \hline
\multirow{9}{3em}{glinast} & 1997 & V soboto , 20. aprila , je bilo na olimpijskem streli\v s\v cu v Ormo\v zu 2. kolo v tretji dr\v zavni ligi v streljanju na glinaste golobe disciplina trap . \newline
(\textbf{eng}.) \textit{On Saturday, April 20, at the Olympic shooting range in Ormo\v z, the 2nd round of the third national league in \underline{clay} pigeon shooting, trap discipline, took place.} & \multirow{9}{3em}{4 4 4} \\
& 2018 & Prekmurka je zadela 114 glinastih golobov , do preboja v veliki finale \v sestih najbolj\v sih strelk na svetu pa sta jo lo\v cila dva zadetka . \newline
(\textbf{eng}.) \textit{The girl from Prekmurje scored 114 \underline{clay} pigeons, only two hits away from reaching the grand final among the six best shooters in the world.} &  \\ \hline
\multirow{12}{3em}{ogaben} & 1997 & Na atletskem stadionu je postavala nepregledna mno\v zica primerno ogabnih brkatih ljubiteljev distorzije , ki pa kak\v snega posebnega zanimanja za Metallico niso kazali . \newline
(\textbf{eng}.) \textit{A huge crowd of pretty \underline{disgusting} mustachioed distortion fans lined up at the athletic stadium, but they didn't show any particular interest in Metallica.} &  \multirow{12}{3em}{4 3 4}\\
& 2018 & Po objavljenih posnetkih so na vrata luksuznih hotelov potrkali policisti in turisti\v cni in\v spektorji , ki so hotelsko upravo kaznovali in pozvali , naj opustijo sramotno in ogabno prakso . \newline
(\textbf{eng}.) \textit{According to the published footage, policemen and tourist inspectors knocked on the doors of luxury hotels, fined the hotel management and called for abandonment of the shameful and \underline{disgusting} practice.}  & \\ \hline
\multirow{8}{3em}{gazela} & 1997 & V zahvalo za meso , ki sem jim ga prepustil , sta dva od njih rada \v sla z menoj ter nesla glavo in stegno gazele . \newline
(\textbf{eng}.) \textit{As a thank you for the meat that I left to them, two of them gladly went with me and carried the head and thigh of a \underline{gazelle}.} & \multirow{8}{3em}{2 2 2}\\
& 2018 & Z veseljem je ugotovil , da nobeno od podjetij , ki se je do danes okitilo z nazivom gazela , ni skrenilo s poti , ampak so-uspela . \newline
(\textbf{eng}.) \textit{He was happy to note that none of the companies that have so far been awarded the \underline{Gazelle} prize have not gone astray, but have succeeded.} & \\

\end{tabular}
\end{table}

\clearpage

\subsection*{Semantic Shift Transport Cost across Model Layers and Regularization Values}
\label{sec:more-results}

\begin{table}[h]
\centering
\caption{Spearman rank correlation for Optimal transport cost using different regularization values ($\lambda$), across model layers and layer combinations. Generally best performing layer ($L_{11}$) is highlighted.}
\label{tab:ot_values}
\begin{tabular}{l|ccccccc}
\hline\hline
Layer & $OT_{\lambda=0.0}$ & $OT_{\lambda=0.05}$ & $OT_{\lambda=0.1}$ & $OT_{\lambda=0.15}$ & $OT_{\lambda=0.2}$ & $OT_{\lambda=0.25}$ & $OT_{\lambda=0.3}$ \\
\hline
$L_0$ & 0.328 & 0.354 & 0.351 & 0.353 & 0.357 & 0.347 & 0.345 \\
$L_1$ & 0.441 & 0.516 & 0.517 & 0.490 & 0.465 & 0.435 & 0.415 \\
$L_2$ & 0.469 & 0.519 & 0.514 & 0.492 & 0.472 & 0.456 & 0.449 \\
$L_3$ & 0.516 & 0.564 & 0.588 & 0.570 & 0.544 & 0.520 & 0.506 \\
$L_4$ & 0.535 & 0.581 & 0.606 & 0.604 & 0.592 & 0.591 & 0.586 \\
$L_5$ & 0.534 & 0.585 & 0.607 & 0.609 & 0.611 & 0.612 & 0.611 \\
$L_6$ & 0.524 & 0.572 & 0.585 & 0.585 & 0.588 & 0.585 & 0.587 \\
$L_7$ & 0.544 & 0.582 & 0.596 & 0.607 & 0.616 & 0.620 & 0.617 \\
$L_8$ & 0.565 & 0.599 & 0.619 & 0.627 & 0.637 & 0.639 & 0.640 \\
$L_9$ & 0.592 & 0.616 & 0.636 & 0.640 & 0.643 & 0.649 & 0.649 \\
$L_{10}$ & 0.630 & 0.643 & 0.669 & 0.683 & 0.687 & 0.686 & 0.687 \\
\cline{2-8}
$L_{11}$ & {0.651} & 0.674 & 0.691 & 0.697 & 0.706 & 0.705 & {0.707} \\
\cline{2-8}
$L_{12}$ & 0.602 & 0.625 & 0.648 & 0.641 & 0.632 & 0.628 & 0.622 \\
$L_{9-11}$ & 0.623 & 0.647 & 0.669 & 0.682 & 0.688 & 0.688 & 0.687 \\
$L_{10-12}$ & 0.632 & 0.655 & 0.679 & 0.692 & 0.695 & 0.699 & 0.697 \\
$L_{9-12}$ & 0.626 & 0.647 & 0.673 & 0.678 & 0.687 & 0.689 & 0.693 \\
\end{tabular}

\begin{tabular}{l|cccccc}
\hline\hline
Layer & $OT_{\lambda=0.35}$ & $OT_{\lambda=0.40}$ & $OT_{\lambda=0.45}$ & $OT_{\lambda=0.50}$ & $OT_{\lambda=0.75}$ & $OT_{\lambda=0.95}$ \\
\hline
$L_0$ & 0.337 & 0.326 & 0.318 & 0.313 & 0.293 & 0.283 \\
$L_{1}$ & 0.395 & 0.383 & 0.378 & 0.372 & 0.363 & 0.354 \\
$L_{2}$ & 0.439 & 0.430 & 0.425 & 0.417 & 0.406 & 0.400 \\
$L_{3}$ & 0.497 & 0.494 & 0.490 & 0.489 & 0.483 & 0.481 \\
$L_{4}$ & 0.580 & 0.573 & 0.571 & 0.569 & 0.570 & 0.568 \\
$L_{5}$ & 0.610 & 0.612 & 0.614 & 0.614 & 0.614 & 0.615 \\
$L_{6}$ & 0.588 & 0.587 & 0.589 & 0.587 & 0.588 & 0.587 \\
$L_{7}$ & 0.619 & 0.619 & 0.618 & 0.619 & 0.621 & 0.622 \\
$L_{8}$ & 0.638 & 0.638 & 0.640 & 0.641 & 0.644 & 0.644 \\
$L_{9}$ & 0.651 & 0.649 & 0.652 & 0.655 & 0.656 & 0.656 \\
$L_{10}$ & 0.688 & 0.690 & 0.688 & 0.689 & 0.691 & 0.692 \\
\cline{2-7}
$L_{11}$ & 0.707 & 0.704 & 0.703 & 0.701 & 0.703 & 0.700 \\
\cline{2-7}
$L_{12}$ & 0.621 & 0.615 & 0.613 & 0.608 & 0.605 & 0.603 \\
$L_{9-11}$ & 0.689 & 0.690 & 0.691 & 0.693 & 0.691 & 0.691 \\
$L_{10-12}$ & 0.697 & 0.697 & 0.696 & 0.695 & 0.695 & 0.697 \\
$L_{9-12}$ & 0.692 & 0.693 & 0.692 & 0.694 & 0.692 & 0.691 \\
\end{tabular}

\end{table}

\clearpage

\printbibliography

\end{document}